\documentclass{article} 
\usepackage{iclr2027_conference,times}

\usepackage{amsmath,amsfonts,bm}

\def\eqref#1{equation~\ref{#1}}

\def\1{\bm{1}}

\DeclareMathAlphabet{\mathsfit}{\encodingdefault}{\sfdefault}{m}{sl}
\SetMathAlphabet{\mathsfit}{bold}{\encodingdefault}{\sfdefault}{bx}{n}

\usepackage{hyperref}
\usepackage{url}
\usepackage{graphicx}
\usepackage{float}
\usepackage{array,booktabs}
\definecolor{gain}{rgb}{0.00,0.42,0.24}
\definecolor{loss}{rgb}{0.70,0.10,0.10}
\usepackage{amsmath,amssymb,amsthm}
\usepackage{titlesec}
\titlespacing*{\section}      {0pt}{8pt plus 2pt minus 2pt}{4pt}
\titlespacing*{\subsection}   {0pt}{7pt plus 2pt minus 2pt}{3pt}
\titlespacing*{\subsubsection}{0pt}{6pt plus 1pt minus 1pt}{2pt}
\titlespacing*{\paragraph}    {0pt}{4pt}{6pt}

\title{Typing Matters: Muscle, Language and \\ Typing Behaviour in Live Keystroke Decoding from sEMG}

\author{%
  Abhilash Durgam\thanks{Correspondence to: \texttt{abhilash.durgam@ucf.edu}} \\
  CRCV, University of Central Florida
  \And
  Nyle Siddiqui \\
  CRCV, University of Central Florida
  \And
  Mubarak Shah \\
  CRCV, University of Central Florida
  \And
  Qiushi Fu \\
  Department of MAE, University of Central Florida
}

\iclrfinalcopy
\begin{document}

\maketitle

\begin{abstract}
Surface electromyography (sEMG) from the wrists can recover typed text without a keyboard, and
is usually posed as a signal-inference problem: characters are read out of the muscle signal and character error rate is reported offline; we argue that this framing removes most of the problem.
The decoder is constrained simultaneously by the muscles, by the language
being typed, and by typing behaviour such as typos, corrections, runs of backspace, line breaks.
A usable system must produce predictions for typed keys live while the hand keeps moving. Existing
systems fall short on two counts: (1) they process sEMG signals in fixed windows and carry nothing
from one window to the next, discarding accumulated context from the typist over a session, and (2) they lack the understanding of the underlying linguistic structure that drives typing. This means
every keystroke is inferred from the muscle signal alone, with little to no ability to leverage any holistic understanding of language. To this end we propose a fully causal,
zero-lookahead sEMG typing decoder built on a state space model (SSM), in which a single model can read and decode a whole sEMG session, without ever requiring windowed partitions, and conditioned on what has already been typed. We achieve this through a novel fusion of a traditional typing decoder with an efficient, autoregressive language model, enabling us to instill both sEMG and linguistic understanding in a singular model. Furthermore, to train such a model  for typing (per-character prediction) rather than whole-world text, we introduce a
retokenization method that rewrites ordinary prose as human-like typing: keystrokes, typos, backspaces and line breaks included. We thus overcome the lack of any large typing corpus by converting the large, text-based WikiText into an accurate, human-like typing dataset. Our model, having never seen a single sEMG-to-character
transcript in typing, achieves a state-of-the-art (SOTA) character error rate of $25.13$\% on emg2qwerty's
cross-user benchmark - a $5.4$\% improvement over the previous SOTA with $6.7\times$ fewer parameters -
while running strictly causally with no future context at all.
\end{abstract}

\section{Introduction}
Turning muscle activity into typed text sits at the intersection of human-computer interaction and
machine learning. Surface electromyography (sEMG) recorded
from a wristband captures the electrical drive to the muscles that move the fingers, so what a person
types can in principle be read from the wrist alone, with no keyboard, no camera, and no line of sight
to the hands. The emg2qwerty dataset~\citep{sivakumar2024emg2qwerty} is the first large-scale
benchmark for this task, with over 346 hours of wrist sEMG from 108 users typing prompted text while
wearing an electrode band on each wrist, and it was motivated by the promise of wrist EMG as an
always-available input channel for AR and VR, for mobile use, and for anyone for whom a physical
keyboard is impractical. The idea is not new: EMG-driven text entry was demonstrated two decades ago
for a forearm amputee~\citep{yu2003emg}, small in scale and accuracy, but enough to establish that muscle
activity alone carries typing information. What has changed is the combination of large datasets and
modern sequence models, and the benchmark has since drawn a fast-moving line of work on encoder
architectures~\citep{sivakumar2024emg2qwerty, hadidi2026splashnet, mehlman2025scaling},
bringing an assistive and general-purpose text interface within reach of a wristband.


\begin{figure*}[t]
  \centering
  \includegraphics[width=\textwidth]{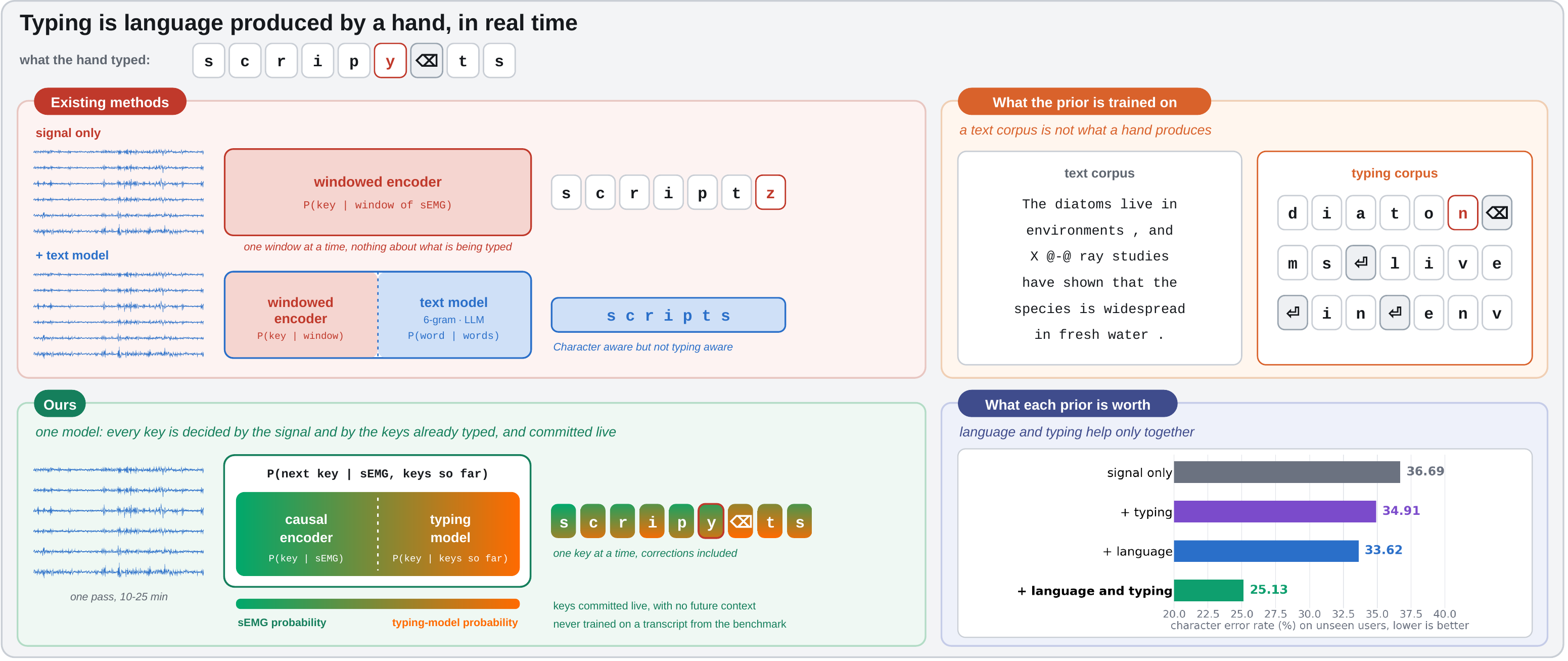}
  \caption{\textbf{Typing is language produced by a hand, in real time.}
  \textit{Left:} existing sEMG decoders map each window of signal to keys on
  its own (\emph{signal only}), or bolt a text model on afterwards that
  re-scores a word once it is finished (\emph{+ text model}). Neither can
  use what has already been typed while the next key is being decided.
  \textit{Ours:} one causal model that commits every key live from
  $P(\text{next key} \mid \text{sEMG}, \text{keys so far})$, combining a
  causal encoder with a per-key typing model; the colour of each output
  key shows how much of that decision came from the signal (blue) versus
  the typing model (red). Corrections such as backspace are ordinary keys.
  \textit{Right:} the typing model is pretrained on text rewritten as
  keystrokes (lowercase, short lines, real corrections) rather than on a
  text corpus, and it is language and typing \emph{together} that lower the
  character error rate on unseen users.}
  \label{fig:teaser}
  \vspace{-.5cm}

\end{figure*}

A lot of effort has gone into the encoder, which leaves the decoder inferring every key
from muscle activity alone. Where a prior over the output is used at all it is a prior over text,
and the benchmark's own baseline is typical of the kind: a character n-gram fitted to encyclopedia
prose, scoring the letters of one word at a time and starting again at every space. But the thing
being decoded is not prose. It is a record of what the hand did, and so it keeps the typos the
typist made, the runs of backspace that undo them and the line breaks between prompts, none of
which occur in written text. A prior built from text has never seen a correction and gives one
almost no probability, and because its context restarts at every space it cannot use what was typed
a word ago. Nor does it know which keys a person confuses, or how long they take to notice a
mistake. What such a prior constrains is the spelling inside a word, and everything else is left to
the signal. The obvious remedy is to fit the prior to typing rather than to text, which runs
straight into the reason nobody has: no corpus of typing exists at the scale a prior needs, because
keystroke logs capture the behaviour but almost none of the language.

An other limitation is structural. Every system on the benchmark reads the signal through a context
of fixed size. The official baseline pairs its convolutional encoder with windows padded by
900\,ms of past and 100\,ms of future context \citep{sivakumar2024emg2qwerty}, SplashNet works from
a receptive field of about a second \citep{hadidi2026splashnet}, and the strongest published system
cuts the session into four-second windows and resets between them \citep{mehlman2025scaling}. A
receptive field at least slides, so what it costs is reach. A window that resets costs reach and
continuity both: four seconds is roughly fifteen keystrokes, so whatever the encoder has established
about this session, and whatever the typist has already written, is thrown away and rebuilt fifteen
times a minute. This is what makes the missing prior hard to repair rather than merely absent,
because a prior that reads across words has nothing to read across once its context is destroyed
every few seconds.

What the limitations cost is clearest against the uses the benchmark \citep{sivakumar2024emg2qwerty} was built for: an input
channel for augmented and virtual reality, for mobile use, and for anyone a keyboard does not
serve. Such a channel has to put each key on the screen as the finger lands, and a four-second
window can leave a key waiting the length of that window before it appears. It has to let a person
repair what they wrote, and a prior fitted to prose has never seen a backspace, so the action a
typist relies on most is the one the decoder least expects. And it has to run all day on a
wristband, which is why the window is there in the first place: read as one unbroken stream, the
attention model behind the best published accuracy needs memory that grows with every frame and
stops keeping pace with the typist after about a minute. The window is not a modelling preference
but the thing that makes such a model runnable, and it is paid for in precisely what an interface
needs.

In this work we introduce a fully causal sEMG typing decoder that reads a whole session as one
unbroken stream. Built on a state-space encoder, it consumes the signal frame by frame and updates
a fixed-size hidden state that compactly summarises everything seen so far, emitting a distribution
over keys at every frame with no lookahead. Unlike windowed approaches it never revisits an input
or restarts; instead it carries that state to the end of the session, so no key after the opening
seconds is decoded from a cold start, at a cost per frame that does not grow. Alongside it we
introduce a typing model trained on keystrokes rather than on text, from a corpus we build by
rewriting ordinary prose as a measured typist would have typed it, with the typos, the backspaces
and the line breaks that a transcript removes. A prefix beam search combines the two at every frame,
so each key is chosen from the muscle signal and from everything already typed at once. The design
follows the thing being decoded: typing is produced left to right, one key at a time, and is undone
only by typing more.

We summarise our contributions as the following:
\begin{itemize}
\item We pose sEMG text entry as decoding typing rather than text: the target is the keystrokes a
hand produced, corrections included, so the decoder needs a prior over typing behaviour and a
context that is not a window.
\item We propose a fully causal state-space decoder that reads a session as one unbroken stream,
where a fixed-size hidden state replaces windowed recomputation at a cost per frame that does not
grow.
\item As an alternative to text priors, we introduce a retokenization that rewrites prose as
keystrokes under an error and correction process fitted to real typists, supplying a typing corpus
where none exists.
\item We report state-of-the-art cross-user accuracy on emg2qwerty, 25.13 against 30.5, strictly
causal and from $6.7\times$ fewer parameters, with a prior that has never seen the benchmark's text.
\end{itemize}

\section{Related Work}
\label{sec:related}

The emg2qwerty benchmark \citep{sivakumar2024emg2qwerty} established the task, the split and the
protocol, and nearly everything published since has been about the encoder: convolutional stacks
have become more efficient \citep{hadidi2026splashnet}, Transformers larger
\citep{mehlman2025scaling}. What these systems share is not how the signal is fed to them but how much
of it they can use. The benchmark's baseline and SplashNet are convolutional, so even when handed a
whole session they see about a second of it at any output frame; the strongest published system is
a transformer over four-second windows and begins each one cold. None carries a learned state from
one part of a session to the next, and the two that report it read some future signal before
committing a key. Every character error rate in this literature, ours among them, is computed once
the recording has finished, so no published number describes what a typist would have seen while
typing. Live EMG interfaces do exist away from this benchmark, with people in the loop, so what is
missing here is not live text entry from muscle but a decoder for these recordings that is causal
and never restarts.

The same work says comparatively little about what is being decoded. Where a prior over the output
appears at all it is a prior over text, and it is the same prior: both the benchmark's baseline and
SplashNet score characters under a character $n$-gram fitted to encyclopaedia prose
\citep{sivakumar2024emg2qwerty,hadidi2026splashnet}, in the shallow-fusion form that CTC speech
recognition made standard \citep{graves2006connectionist,hannun2014deep}. Both searches are
backspace-aware, and aware in the same precise sense: when a backspace arrives the search retracts
the language model's last contribution, so that a typo and its correction score as though neither
had happened. The language model itself is never asked about a backspace. A correction therefore
contributes no evidence, cannot be anticipated, and is carried entirely by the muscle signal. We
know of no decoder in this literature whose prior is fitted to keystrokes rather than to text.

Fitting a prior to keystrokes instead would need a corpus of them, and the two ingredients such a
corpus requires have never been available together. Keystroke logs hold the behaviour and almost
none of the language: Aalto 136M \citep{dhakal2018observations} records every keypress of
$168{,}593$ volunteers, who between them typed only $1{,}525$ distinct sentences. Prose corpora
\citep{merity2017pointer} hold the reverse, all of the language and none of the behaviour, with no
typo, no backspace and no line break of the kind a real session contains. Nothing public holds both
at the scale a prior needs, which is why this work builds such a corpus rather than finding one.

Turning text into something that looks mistyped is not itself new. Spelling-error channels have
been estimated from real misspellings for decades
\citep{kernighan1990spelling,brill2000improved}, and synthetic noise is routinely injected to make
models robust or to train grammatical error correction
\citep{belinkov2018synthetic,karpukhin2019training,grundkiewicz2019neural,zhuang2021robustly}. None
of it emits a correction. These methods produce text that contains errors rather than a stream of
keystrokes in which an error is noticed and then deleted, so the runs of backspace, the delay before
a typist reacts and the errors that are never repaired at all do not occur in what they generate.
Correction is modelled instead by simulated typists built for touchscreens
\citep{shi2024crtypist,shi2025typoist}, where the unit is a tap coordinate and the model includes
where the eyes are looking, and those exist to evaluate interfaces rather than to train a prior over
physical keys.

\section{Methodology}
\label{sec:method}

\begin{figure*}[t]
  \centering
  \includegraphics[width=\textwidth]{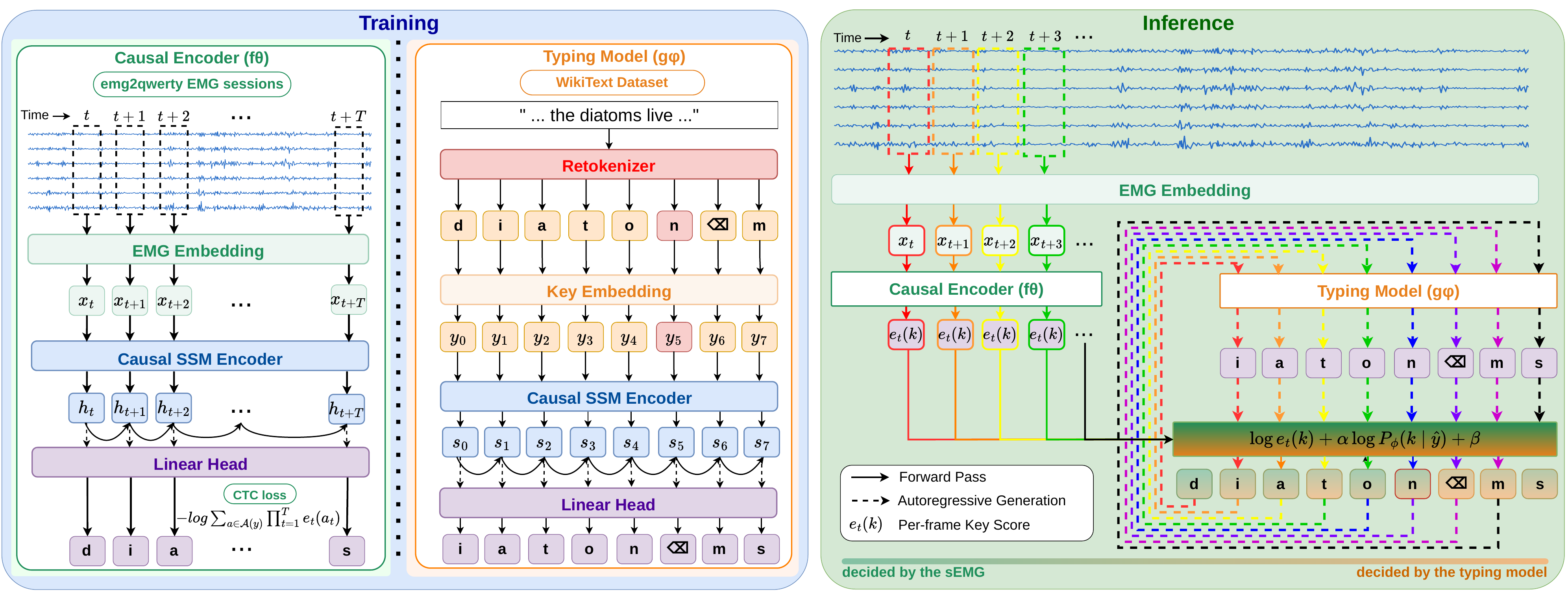}
  \caption{\textbf{Training and inference.} \textit{Left:} the two trained parts, each shown as its
inputs, its recurrence and its outputs. The causal encoder $f_\theta$ reads sEMG frames from
emg2qwerty's training sessions, embeds them, advances a state $\mathbf{h}$ one frame at a time and
emits a key-or-blank distribution $e_t(k)$ at every frame; it is trained with CTC. The typing model
$g_\phi$ reads a line of WikiText rewritten as keystrokes by the retokenizer, in  the figure \texttt{diaton}
mistyped and repaired to \texttt{diatom}, advances a state $\mathbf{s}$ one key at a time and
predicts the next key; it is trained by cross-entropy and never sees emg2qwerty text.
\textit{Right:} at inference a single causal pass reads the session with no windows and no resets.
The encoder scores every key at every frame, the typing model scores the same keys given the prefix
decoded so far, and the beam adds the two scores. The gradient shows how much of each key's decision came from the sEMG rather than the typing model.}
  \label{fig:model}
\end{figure*}

We present a causal sEMG typing decoder (Fig.~\ref{fig:model}) for continuous streaming inference.
Unlike prior approaches that partition a session into fixed windows and decode each one independently,
our method continuously reads a session forward in a single pass, enabling it to model and predict typing behaviors driven by linguistic understanding rather than purely from sEMG signals alone. Our method is built on two key principles:
(1) causal, linear-time sequence modeling to enable continuous keystroke streaming at a constant cost that does not grow with the session; and (2) re-framing autoregressive language modeling for human-like typing rather than whole-word text such that special characters like typos, backspaces, and line breaks are events the understood and predictable by the  decoder. We first provide some background on sEMG signals and state space models
(Section~\ref{sec:prelim}), followed by the three parts of the system: (i) the causal encoder, (ii)
the typing model, and (iii) our proposed search that combines predictions from both models (Section~\ref{sec:model}), and finally the custom-built typing dataset we propose to train such a model (Section~\ref{sec:retok}).

\subsection{Preliminaries}
\label{sec:prelim}

\paragraph{sEMG framing.}
We represent a session as a multichannel time series $\mathbf{X}\in\mathbb{R}^{C\times T}$, where
$C$ denotes the number of electrodes across the two wristbands and $T$ the number of temporal
samples. Rather than operating directly on raw samples, each electrode's signal is turned into a
log-spectrogram, its bins are pooled into a few broad bands, and the two hands are joined into one
vector per frame, yielding feature frames $\mathbf{x}_{1:T'}$ where each
$\mathbf{x}_t\in\mathbb{R}^{d}$ summarizes the signal at temporal index $t$ across all electrodes.
Unlike patch-based encoders, we neither group frames into non-overlapping patches nor downsample in
time, so a key can be emitted at the frame it is pressed. The ground truth pressed keys, $y=y_{1:L}$, are drawn from the key set $\mathcal{K}$ proposed in \cite{sivakumar2024emg2qwerty}: all alphabetic letters combined with special characters such as space, backspace, enter, shift, etc. The key difference in typing data stems from these special characters as well as the mistakes that humans make during typing - backspaces and spelling corrections must also be predicted by a typing model when reading sEMG signals. Moreover, we define the full key set as $\bar{\mathcal{K}}=\mathcal{K}\cup\{\varepsilon\}$, which includes a CTC blank character $\varepsilon$ which allows the model to denote that no key was pressed during a specific window.

\paragraph{State space models.}
We leverage an SSM to encode sEMG signals, which treats a continuous sequence as the evolution of a latent
dynamical system, operates in near-linear complexity with respect to sequence length, and excels at capturing
long-range dependencies. Given an input sequence, SSMs maintain a latent state
$\mathbf{h}_t\in\mathbb{R}^{N}$ that evolves over time and retains only what it deems as salient information for the given training objective. To discretize these continuous operations, a learnable step size parameter $\Delta$, is applied and yields:
\begin{equation}
  \mathbf{h}_t = \bar{A}\,\mathbf{h}_{t-1} + \bar{B}\,\mathbf{x}_t,
  \qquad
  \mathbf{y}_t = \bar{C}\,\mathbf{h}_t ,
  \label{eq:ssm}
\end{equation}
where $\bar{A}$ governs state transitions and $\bar{B}$ and $\bar{C}$ denote input and output
projections respectively. SSMs can be parameterized as either a global convolution or as a recurrence, thus allowing us to parallely train using the convolutional parameterization while using the recurrent parameterization during inference for online, linear-time scaling. Specifically, we use a selective state space model \citep{lahoti2026mamba3}, in which $\Delta$, and through it $\bar{A}$ and $\bar{B}$, are input-dependent functions
of $\mathbf{x}_t$. Selectivity gives up the convolutional form, and parallel evaluation becomes a
scan, but it is what allows the model to decide at every frame how much of the state to keep and how
much to write, which Appendix~\ref{app:internals} finds is what the encoder depends on most.

This makes SSMs a prime candidate for live, full session sEMG decoding, as the state at $t$ depends only on $\mathbf{x}_{1:t}$, which is what live decoding requires, and
inference advances one frame at a time at constant memory and compute cost per frame. We provide additional details on this in Appendix~\ref{app:cost}.
Throughout, $W$ with a subscript denotes a learned linear projection, $E$ a learned embedding
matrix, and $\gamma,\beta$ a learned per-feature affine; $P$ is a fixed pooling matrix and
$\epsilon$ a small constant.

\vspace{-.25cm}



\subsection{The Model}
\label{sec:model}

We provide an overview of the model in Figure~\ref{fig:model}, and subsequently describe each component: from the raw sEMG session $\mathbf{X}$ to the 
decoded keystrokes.

\paragraph{sEMG embedding.}
To perform live, casual sEMG decoding, an embedding must not look ahead at future embeddings and must not be constrained to a simple sliding window approach . Writing $\mathbf{X}_c$ for the signal of electrode $c$, each electrode is turned into
a log-spectrogram whose bins are pooled by a fixed matrix $P$ into $B$ broad bands; each frame is
then normalised by the running mean $\mu_t$ and variance $\sigma^2_t$ of frames $1$ to $t$; the
electrodes of each hand $h$ pass through a rotation-invariant MLP \citep{sivakumar2024emg2qwerty} that averages over the rotations
$\mathcal{R}$ of the electrode ring; and the two hands are joined into one frame:
\begin{equation}
\begin{aligned}
  \mathbf{u}_{c,t} &= P\,\log\big|\mathrm{STFT}(\mathbf{X}_c)_t\big|^{2},
  &\quad
  \mathbf{r}^{(h)}_t &= \frac{1}{|\mathcal{R}|}\sum_{\rho\in\mathcal{R}}
    \mathrm{MLP}\big(\rho\cdot\mathbf{v}^{(h)}_t\big), \\[2pt]
  \mathbf{v}_{c,t} &= \gamma \odot
    \frac{\mathbf{u}_{c,t}-\mu_t}{\sqrt{\sigma^{2}_t+\epsilon}} + \beta,
  &\quad
  \mathbf{x}_t &= \mathrm{GELU}\big(W\,\mathrm{LN}(
    [\,\mathbf{r}^{(\mathrm{L})}_t;\ \mathbf{r}^{(\mathrm{R})}_t\,])\big)
    \in \mathbb{R}^{d}.
\end{aligned}
\label{eq:emb}
\end{equation}
Note that the normalization and accumulated sEMG statistics for a given session are strictly causal and do not look ahead of frame $t$. 

\paragraph{Causal SSM Encoder.}
To ensure both causality and a constant cost across a minutes-long sEMG session, our encoder is a stack of $L$ selective state space blocks
\citep{lahoti2026mamba3}, each a pre-normalised residual around the recurrence of
Equation~\ref{eq:ssm} whose $\Delta$, and through it $\bar{A}$ and $\bar{B}$, are produced from
its own input:
\begin{equation}
  \mathbf{x}^{(\ell)}_t = \mathbf{x}^{(\ell-1)}_t
    + \mathrm{SSM}^{(\ell)}\!\big(\mathrm{RMSNorm}(\mathbf{x}^{(\ell-1)})\big)_t ,
  \qquad \mathbf{x}^{(0)} = \mathbf{x}, \qquad \ell = 1,\dots,L.
  \label{eq:enc}
\end{equation}
A linear head reads the last block out at every frame, with no downsampling to delay a key:
\begin{equation}
  e_t(k) \;=\; P_\theta\big(k \mid \mathbf{x}_{1:t}\big)
  \;=\; \mathrm{softmax}\big(W_e\,\mathrm{RMSNorm}(\mathbf{x}^{(L)}_t)\big)_k ,
  \qquad k \in \bar{\mathcal{K}} .
  \label{eq:emit}
\end{equation}
Since $\mathbf{x}_t$ depends on $\mathbf{X}$ up to $t$ and every block is a forward recurrence, the
output at $t$ depends on $\mathbf{x}_{1:t}$ alone, as live decoding needs, and a frame costs the
same at the end of a session as at its start, as carrying the session needs. The encoder is trained
with connectionist temporal classification \citep{graves2006connectionist},
\begin{equation}
  \mathcal{L}_{\mathrm{CTC}}(\theta) \;=\; -\log \sum_{\mathbf{a}\in\mathcal{A}(y)}
    \;\prod_{t=1}^{T} e_t(a_t),
  \label{eq:ctc}
\end{equation}
where $\mathcal{A}(y)$ is the set of frame-by-frame labellings $\mathbf{a}\in\bar{\mathcal{K}}^{T}$
that spell $y$ once their blanks are dropped and a key held over consecutive frames is counted once;
for example, with $y=\mathtt{it}$ over five frames,
$\varepsilon\,\mathtt{i}\,\mathtt{i}\,\varepsilon\,\mathtt{t}$ and
$\mathtt{i}\,\varepsilon\,\varepsilon\,\mathtt{t}\,\mathtt{t}$ are both in $\mathcal{A}(y)$.

\paragraph{Typing model.}
While most previous works follow a similar architecture up until this point, a crucial difference is the disconnect between simple next-key prediction and actual linguistic understanding. While the former has already been largely explored, we hypothesize leveraging the latter will generally improve performance due to the entangled nature of typing and language. 
Existing decoders
use a text prior over prose tokenised into words: it scores the characters of one word at a time,
its context reset at every space, and prose has no backspace in it
\citep{sivakumar2024emg2qwerty,hadidi2026splashnet}. Ours tokenises typing into keys, one token per
key pressed, backspace and enter included, and reads across words. The typing model $g_\phi$ is a
second stack of the same blocks, reading keys where the encoder reads frames: a key embedding $E$
carries $y_{i-1}$ into a state $\mathbf{s}_i$ advanced once per decoded key, and a linear head gives
the next-key distribution,
\begin{equation}
  \mathbf{s}_i = \mathrm{SSM}_\phi\big(\mathbf{s}_{i-1},\, E\,y_{i-1}\big),
  \qquad
  P_\phi\big(y_i \mid y_{<i}\big) = \mathrm{softmax}\big(W_\phi\,\mathrm{RMSNorm}(\mathbf{s}_i)\big).
  \label{eq:tm}
\end{equation}
Training such a model requires next-key supervision over realistic human typing, rather than conventional text alone. Large-scale text corpora provide rich linguistic structure, but they contain only the final written text and omit the keystrokes that produced it - including typos, backspaces, corrections, and other special keys. We therefore train the typing model with next-key cross-entropy on our own curated typing corpus introduced in Section~\ref{sec:retok}. Because the typing model shares the encoder's alphabet $\bar{K}$, a backspace is treated as an ordinary token, allowing correction behavior to be learned directly from the data rather than imposed through a hand-designed rule or statistic. At decode time, the model conditions only on the most recent $K$ keys, roughly ten words, keeping inference inexpensive enough to evaluate at every frame. 

\paragraph{Decoding.}
The search must commit the keys that were pressed, corrections included, while the hand is still
moving. A CTC prefix beam search \citep{hannun2014deep} combines the two models. Each hypothesis is
a prefix $\hat{y}$ of decoded keys with its blank-ending and key-ending probabilities under the
emissions; at frame $t$, extending $\hat{y}$ by a key $k$ adds
\begin{equation}
  \log e_t(k) \;+\; \alpha\,\log P_\phi\big(k\mid \hat{y}\big) \;+\; \beta_{\mathrm{ins}} ,
  \label{eq:fusion}
\end{equation}
where $\alpha$ weights the typing model and $\beta_{\mathrm{ins}}$ is a per-key insertion bonus,
both chosen on validation sessions, and only the best-scoring hypotheses survive each frame. Both
terms are causal: $e_t$ uses only $\mathbf{x}_{1:t}$, and $P_\phi$ only the keys already in the
prefix. 

\paragraph{Training.}
The causal encoder is trained on the emg2qwerty training sessions using a three-stage curriculum, with each stage initialized from the preceding checkpoint. We first train on short windows, where the model learns the majority of the local sEMG-to-keystroke mapping. We then extend training to windows spanning several minutes, encouraging the recurrent state to retain useful information over substantially longer temporal contexts. Finally, we perform session-stateful training, in which each session is processed sequentially while the recurrent SSM state and causal normalization statistics are carried across chunk boundaries and detached between chunks. This final stage exposes the model to the same persistent-state setting used at inference, allowing the internal state to evolve over a full session without periodic resets. The intermediate long-window stage is necessary to stabilize this behavior, as directly transitioning from short-window training to reset-free inference causes the encoder's emission rate to drift. The typing model is trained independently using only the corpus introduced in Section~\ref{sec:retok}. Appendix~\ref{app:hparams} provides the complete training configuration, decoding constants, and details on which hyperparameters were selected using validation data.
\vspace{-.25cm}

\subsection{Rewriting text as typing}
\label{sec:retok}

Typing is not text with noise sprinkled over it. Who is typing, which words are difficult, which keys
are likely to be confused, and when an error is noticed and corrected all follow structured patterns
of their own. A prior over typing must therefore model these behaviors in addition to the underlying
language.

The typing model of Section~\ref{sec:model} operates directly over keystroke sequences, including
backspace and enter, and therefore requires next-key supervision over realistic human typing.
However, no typing corpus exists at the scale needed to train such a model. Keystroke logs record
typing behaviour but contain limited linguistic diversity: in Aalto 136M
\citep{dhakal2018observations}, $168{,}593$ volunteers copy-typed sentences with every keystroke
logged, yet only $1{,}525$ distinct sentences were typed. Prose has the opposite problem: it provides
large-scale linguistic variation but contains only finalized text, with no record of the errors,
corrections, or special keys that produced it. We combine these complementary sources by
retokenizing prose as typing.
With $u$ a line of WikiText-103 \citep{merity2017pointer} drawn from $P_{\mathrm{txt}}$ and
$Q_\psi(y\mid u)$ a retokenizer that converts it into a human-like keystroke sequence, the typing
corpus is drawn from
\begin{equation}
  \textstyle\sum_u P_{\mathrm{txt}}(u)\,Q_\psi(y\mid u),
  \label{eq:retok}
\end{equation}
so language enters through $u$ and typing behaviour through $\psi$. Every parameter of $\psi$ is
fitted on Aalto: replaying each keystroke stream against its sentence
\citep{wobbrock2006analyzing} recovers its errors, corrections, and deletions for the $151{,}278$
typists who yield a posterior, with $7{,}875$ held out for validation. The simplest retokenizer, and the one we replace, errs at a single
rate on every key and corrects every error. Ours turns a line $u$ into a keystroke stream $y$ in
four draws: who is typing, which words go wrong, which key is hit instead, and what happens next.

\paragraph{Who is typing.}
Typists differ, and a session belongs to one of them, so the prior should see one person's habits
throughout a session rather than an average typist at every key. Each synthetic session draws a real
typist $j$ from the Aalto pool and takes three things from their record: how often they err,
$\zeta_j$; how quickly they notice an error, $\xi_j$; and how far back they delete, $\varsigma_j$,
each drawn from that typist's fitted posterior. This produces sessions with internally consistent
typing behaviour, rather than resampling population-level tendencies at every key.

\paragraph{Which words go wrong.}
Errors are not distributed uniformly across words: long, rare, and mechanically difficult words are
more likely to be mistyped. We assign each word $w$ an error probability from three features
$r_w$: its length, how common it is, and how awkward its key pairs are,
\begin{equation}
  p_w=\sigma\!\left(\omega_0+\omega^{\top}r_w+\log m_w\right),
  \label{eq:retok-risk}
\end{equation}
where $m_w$ adjusts common words on their own record, and the weights $\omega$ are fitted on Aalto
and transferred to Wikipedia's vocabulary by importance weighting
\citep{shimodaira2000improving}. The typist determines how many words contain errors,
$n\sim\mathrm{Poisson}(\zeta_j\sum_w p_w)$, while the text determines which words are selected:
$n$ words are drawn with inclusion probability proportional to $p_w$ using Sampford's sampling
design \citep{sampford1967sampling}.

\paragraph{Which key is hit.}
The identity of the incorrect key also matters, since the typing model should not assign the same
likelihood to \texttt{teh} and \texttt{tqh}. Within an erring word, the error position is drawn from
a base law tilted toward difficult key pairs, and the incorrect key is sampled from a confusion
kernel $C(k\mid u_{i-1},u_i,u_{i+1})$ that backs off to shorter contexts. This encourages errors
that reflect plausible local key confusions rather than arbitrary substitutions. In some cases the
entire word $v$ becomes another valid word $v'$, with probability decreasing with edit distance
$d(v,v')$ and increasing with frequency $f(v')$. These real-word errors are particularly important
because a conventional spelling prior may assign them high probability despite the underlying
typing mistake.

\paragraph{What happens next.}
A prior trained only on prose provides no model of what happens after an error is made. In real
typing, a correction may occur several keys later, may require a run of backspaces, or may never
occur at all. After each further
key an open error is noticed with hazard $\nu=\sigma(\chi^{\top}z+\xi_j)$, where $z$ records how many
keys ago it happened and whether a word boundary has just been crossed. A noticed error is deleted
back to the error, to the start of the word or too far, the target drawn from a lag table tilted by
$\varsigma_j$, and the deleted text is retyped and may err again. What leaves the retokenizer is
$y$, a sequence over the same alphabet $\mathcal{K}$ the encoder emits, with the typos, the
backspace runs and the line breaks in place.

Nothing in $Q_\psi$ is fitted on the benchmark: the error process, line lengths, and lines per session
all come from Aalto, and five global intercepts are tuned so that the sampler, when applied to the
held-out typists' prompts, matches their measured rates. The prompts of emg2qwerty
\citep{sivakumar2024emg2qwerty} also include random word lists, which the corpus does not attempt
to imitate. No benchmark statistic enters $\psi$, and no benchmark transcript enters the corpus.
\vspace{-.25cm}
\section{Experiments}
\label{sec:experiments}
\vspace{-.25cm}

\subsection{Datasets and evaluation protocol}
\label{sec:protocol}

We follow the official emg2qwerty protocol \citep{sivakumar2024emg2qwerty}. The dataset holds 1,135 touch-typing sessions from 108 users, 346 hours in all, recorded at 2\,kHz from two 16-electrode wristbands while participants copied prompted word lists and Wikipedia sentences. The ground truth is the keylogger record of what the hand typed, backspace and enter included, so a typo and its correction are both part of the target.

The generic split, as released, trains on 837 sessions and validates on 192 sessions from 96 users, and tests on 16 sessions from 8 held-out users who appear in neither. Training uses short windows cut from each session; at test time every session is fed to the model in its entirety, without windowing or padding. The metric is character error rate (CER): the Levenshtein edit distance between the decoded and the ground-truth key sequences, with backspace and enter as literal characters, divided by the ground-truth length and pooled over all keys of the 16 test sessions. The official baseline pairs a TDS convolutional CTC encoder with either greedy decoding or a backspace-aware beam search under a character 6-gram trained on WikiText-103.

We use the released split unchanged, decode every test session whole, and score with a metric checked to be identical to the official scorer; decoding constants were selected on validation sessions of training users (Appendix~\ref{app:hparams}).
\vspace{-.5cm}

\subsection{Results}
\label{sec:results}
\vspace{-.25cm}

Table~\ref{tab:main} lists every published cross-user result on the benchmark under its own design, and ours. The
previous published best is the Large model of \citep{mehlman2025scaling} at 30.5, decoded inside 4\,s windows with 100\,ms of
future context. Ours is 25.13: each session decoded whole in one causal pass, with no future context, no window
boundaries, no state reset, and a typing model that has seen no emg2qwerty text, from 16.24M parameters in total,
6.7 times fewer than their encoder alone. The two rows are not the same experiment, and the stricter protocol is
ours; it still finishes 5.37 CER below the best published number.

\begin{table}[H]
\centering\small
\setlength{\tabcolsep}{5pt}
\caption{Cross-user CER (\%) on the emg2qwerty generic test set (16 whole sessions, 8 unseen users, 5 seed runs).}
\label{tab:main}
\begin{tabular}{@{}lllr@{}}
\toprule
System & Encoder & Params & CER ($\downarrow$) \\
\midrule
emg2qwerty, greedy \citep{sivakumar2024emg2qwerty} & TDS conv & 5.29M & $55.38\pm4.10$ \\
emg2qwerty, beam + char.\ 6-gram & TDS conv & 5.29M & $51.78\pm4.61$ \\
splashnet-mini \citep{hadidi2026splashnet} & conv & 1.38M & $36.41\pm7.30$ \\
splashnet & conv & 2.58M & $35.67\pm6.79$ \\
Mehlman et al.\ Small \citep{mehlman2025scaling} & Transformer & 5.4M & $35.2\pm1.1$ \\
Mehlman et al.\ Tiny, distilled from Large & Transformer & 2.2M & $31.9\pm0.4$ \\
Mehlman et al.\ Large & Transformer & 109M & $30.5\pm0.6$ \\
\midrule
Mehlman et al.\ Large + our typing model & Transformer & 109.54M + 8.18M & $26.77\pm0.43$ \\
Ours: causal encoder + typing model & Mamba & 8.06M + 8.18M & $\mathbf{25.13}\pm0.51$ \\
\bottomrule
\end{tabular}
\vspace{-.25cm}

\end{table}

Where that difference comes from can be tested directly, by giving the prior away. Placed on their Large encoder,
inside their own 4\,s windows and their own protocol, our typing model produces the largest single improvement in
the table, from 30.5 to 26.77 and still finishes 1.64 above our own system (95\% CI 0.86--2.42, $p<0.01$). What
remains is the window. Their decoder sees 4\,s at a time and starts each one cold, so the encoder's state restart about every fifteen keystrokes, while ours run to the end of the session. A prior over typing can only use what it is allowed to remember.

\vspace{-.25cm}

\subsection{Ablations}
\label{sec:ablations}

\subsubsection{What each part is worth}
\label{sec:abl-ladder}

We measure what the priors are worth by giving three of them the same network and the
same search, so they differ only in what they read. Measured keystrokes alone are worth
1.78 CER over the greedy readout and prose alone 3.07, while prose rewritten as typing is
worth 11.56, more than twice what the two parts add up to (Table~\ref{tab:abl-ladder}).

\subsubsection{What the rewriting is worth}
\label{sec:abl-retok-cer}

To measure the effectiveness of our typing sampler we have measured a rewriting that errs at random,
which has the same errors and backspaces as ours but none of the realism. It costs 3.33 CER
(Table~\ref{tab:retok-cer}), so the errors have to look like real typos and not merely be there.

\vspace{-.25cm}

\begin{table}[H]
\centering\small
\begin{minipage}[t]{0.48\textwidth}
\centering
\caption{What the prior is trained on (test set, CER \%).}
\label{tab:abl-ladder}
\setlength{\tabcolsep}{4pt}
\begin{tabular}{@{}lrr@{}}
\toprule
Prior trained on & CER & \\
\midrule
greedy readout & 36.69 & \\
keystrokes & 34.91 & \textcolor{gain}{$-1.78$} \\
prose & 33.62 & \textcolor{gain}{$-3.07$} \\
rewritten as typing & 25.13 & \textcolor{gain}{$-11.56$} \\
\bottomrule
\end{tabular}
\end{minipage}\hfill
\begin{minipage}[t]{0.48\textwidth}
\centering
\caption{Priors trained on differently rewritten corpora (test set, CER \%).}
\label{tab:retok-cer}
\setlength{\tabcolsep}{4pt}
\begin{tabular}{@{}lrr@{}}
\toprule
Prior trained on & CER & \\
\midrule
greedy readout & 36.69 & \textcolor{loss}{$+11.56$} \\
random errors & 28.46 & \textcolor{loss}{$+3.33$} \\
ours & 25.13 & \\
\bottomrule
\end{tabular}
\end{minipage}

\par\vspace{10pt}

\begin{minipage}[t]{0.40\textwidth}
\centering
\caption{Cost of resetting the decoder's state at a fixed period (test set, CER \%).}
\label{tab:abl-resetperiod}
\setlength{\tabcolsep}{4pt}
\begin{tabular}{@{}lrr@{}}
\toprule
Reset period & CER & \\
\midrule
never (as shipped) & 25.13 & \\
every 256\,s & 25.42 & \textcolor{loss}{$+0.29$} \\
every 32\,s & 28.33 & \textcolor{loss}{$+3.20$} \\
\bottomrule
\end{tabular}
\end{minipage}\hfill
\begin{minipage}[t]{0.57\textwidth}
\centering
\caption{Autocomplete on held-out prose (\%). Completion is top-1 with three letters typed; savings
is keystrokes not pressed, over 1,200 lines.}
\label{tab:autocomplete}
\setlength{\tabcolsep}{4pt}
\begin{tabular}{@{}lrrr@{}}
\toprule
 & & \multicolumn{2}{c}{Savings} \\
\cmidrule(l){3-4}
Typing model & Compl. & top-1 & top-3 \\
\midrule
character 6-gram & 37.2 & 22.3 & 35.3 \\
\textbf{ours, 8.18M} & \textbf{69.2} & \textbf{43.3} & \textbf{54.3} \\
\textbf{ours, 61.32M} & \textbf{75.1} & \textbf{47.3} & \textbf{58.0} \\
\bottomrule
\end{tabular}
\end{minipage}
\end{table}

\subsubsection{Session state}
\label{sec:abl-state}

We ablate the decoder's memory by resetting it at a fixed period, the way a windowed decoder does.
State lives in two places: the encoder's recurrent state over the signal, and the typing model's
history of the keys already decoded. A window resets both at the same instant, so the damage
concentrates at the boundaries, and it grows as the windows get shorter
(Table~\ref{tab:abl-resetperiod}). Cutting the typing model's history on its own, from sixty-four
keys to eight, is worth 2.5 CER.

\subsubsection{Autocomplete}
\label{sec:abl-autocomplete}

Since the typing model is trained autoregressively, it can be asked for a key before that key is pressed, so we
ablate its effectiveness at autocomplete as an application. On held-out prose it offers the intended word as its top
suggestion after three letters about two times in three, and with a single suggestion it removes 43.3\% of the
keystrokes that would otherwise be pressed, against 22.3\% for the character 6-gram the benchmark ships. We also
wanted to see what happens when the typing model is scaled, so we trained one seven times larger; every measure
improves (Table~\ref{tab:autocomplete}). 

\section*{AI Use Statement}
In this work, we used generative AI tools only to assist with writing and grammar in the manuscript text.
We have not used generative AI tools for
generating synthetic data, developing theoretical models or conceptual frameworks, formulating or proving mathematical claims,
proposing or refining hypotheses, designing or giving feedback on research methodology or experiments, implementing methods,
translation, cleaning or reformatting datasets, qualitative or thematic data analysis, or interpreting results.
We have reviewed all AI-assisted text, and we take responsibility for the final content of this work,
including text, claims, or artifacts produced with the aid of generative AI.

\section*{Ethics Statement}
This work uses only publicly released, de-identified corpora: the emg2qwerty wrist sEMG
benchmark \citep{sivakumar2024emg2qwerty}, the Aalto keystroke logs used to fit the typing
sampler \citep{dhakal2018observations}, and WikiText-103 \citep{merity2017pointer}, each
used under its own terms. No new human-subject data were collected. A wrist-worn decoder
that recovers keystrokes recovers everything typed, including text a person never intends
to send, so deployment of such an interface should keep decoding on the device and under
the wearer's control. The system reported here is a research prototype and has not been
evaluated with the populations the benchmark was motivated by.

\section*{Reproducibility Statement}
The evaluation protocol, including the benchmark's user-disjoint split and the whole-session
decoding condition, is given in Section~\ref{sec:protocol}. The keystroke rewriting that
produces the typing corpus is specified in Section~\ref{sec:retok}. Architecture, the
optimisation settings for both training phases and for the typing model, and the decoding
constants, together with the validation evidence behind them and which were inherited rather
than tuned, are given in Appendix~\ref{app:hparams}.

\bibliography{iclr2027_conference}
\bibliographystyle{iclr2027_conference}

\appendix
\section{Appendix}

\subsection{Implementation details}
\label{app:hparams}

\paragraph{EMG embedding.}
Two wristbands, sixteen electrodes each, sampled at 2\,kHz. Every electrode is turned into a
log-spectrogram with a 64-sample window (32\,ms) and a hop of 16 samples, giving 125 frames per
second and 33 bins, and the bins are sum-pooled into six fixed bands with edges at 0, 62.5, 125,
250, 375, 687.5 and 1000\,Hz. Each feature is then normalised in fp32 by the cumulative mean and
variance of frames $1$ to $t$, with a learnable per-feature affine and a warm-up in which the first
125 frames share one set of statistics. A rotation-invariant MLP maps 96 to 384 per hand, averaged
over electrode-ring rotations $\{-1,0,+1\}$, and the two hands are joined by
LayerNorm(768)\,$\to$\,Linear(768,384)\,$\to$\,GELU. During training only, two thirds of minibatches
have 55\% of feature columns zeroed for the whole sample, and we apply band rotation, alignment
jitter up to 60\,ms and SpecAugment with three time masks and two frequency masks.

\paragraph{Encoder.}
Eight Mamba blocks at $d_{\mathrm{model}}=384$, $d_{\mathrm{state}}=64$, expand 2, head dimension
64 giving twelve heads, pre-RMSNorm residuals and a final RMSNorm, with residual dropout 0.1 during
training. A linear head maps 384 to 99 outputs, the 98 keys plus the CTC blank, at the full 125\,Hz
with no temporal downsampling so that a key can be emitted at its own frame. An auxiliary CTC head
after block four carries loss weight 0.5 against 1.0 for the top head and is discarded at
inference. The encoder is 8.06M parameters at inference.

\paragraph{Encoder training.}
AdamW throughout, weight decay $10^{-4}$, gradient clipping 1.0, bfloat16 autocast with the running
normalisation kept in fp32. Phase A is 30 epochs on 32\,s windows at learning rate
$5\times10^{-4}$, cosine annealed to a floor of $0.1\times$, with 32 windows per optimiser step.
Phase B continues for 6 epochs on 256\,s windows at the annealed floor, again 32 windows per step,
carrying the optimiser state. Phase C is a session-stateful continuation for 3 epochs at learning
rate $5\times10^{-5}$ after 100 warm-up steps: sixteen sessions are held open at once, each
advanced by one 32{,}000-frame chunk per step, with the recurrent, normalisation and rotary states
carried across chunk boundaries and detached between them. The first chunk of each session is drawn
uniformly between 4{,}000 frames and the full chunk length so that boundaries do not align across
epochs, every chunk length is a multiple of 64, and the typing model is frozen. The control that
establishes the carried state is the active ingredient repeats Phase C with every chunk started
cold.

\paragraph{Typing model.}
Embedding($99\times384$), eight Mamba blocks at $d_{\mathrm{model}}=384$, $d_{\mathrm{state}}=128$,
expand 2, head dimension 64 and no dropout, then RMSNorm and Linear(384,99); 8.18M parameters. It
shares the encoder's alphabet, so backspace, enter and shift are ordinary tokens, and
begin-of-sequence is the newline key. Training is next-key cross-entropy on 256-key segments with
AdamW, one-cycle learning rate $3\times10^{-3}$ with 10\% warm-up, weight decay 0.1, batch 64,
gradient clipping 1.0 and bfloat16, for 8 epochs  over a corpus of 538{,}219{,}404 keys. The
checkpoint is the best bits per key on the corpus's own validation split, which contains no
benchmark text.

\paragraph{Hardware.}
Phase A and B on two H100 PCIe GPUs, Phase C and every decode on a single H100 PCIe.

\subsection{Cost of decoding a whole session}
\label{app:cost}

Our encoder carries a fixed-size recurrent state, so the memory it needs to decode does not depend
on how long the session has run. A Transformer doing the same thing must keep a key-value cache
over every frame it has seen, which grows without bound (Table~\ref{tab:app-memory}). This is why
window-based systems window: at twenty minutes the cache is over two thousand times the size of our
state. Both columns are exact, computed from the architectures at 125 frames per second in
bfloat16: eight layers of a 768-wide inner dimension with 64 state channels for ours, against eight
layers of 384-wide keys and values per frame.

\begin{table}[H]
\centering\small
\setlength{\tabcolsep}{6pt}
\caption{Memory carried while decoding, by session length (MB). Ours is a recurrent state; the
Transformer is a parameter-matched causal stack of the same depth and width, keeping a key-value
cache. Weights are excluded from both, being identical and constant.}
\label{tab:app-memory}
\begin{tabular}{@{}lrr@{}}
\toprule
Session length & Ours & Transformer \\
\midrule
30\,s   & \textbf{0.79} & 46 \\
1\,min  & \textbf{0.79} & 92 \\
2\,min  & \textbf{0.79} & 184 \\
5\,min  & \textbf{0.79} & 461 \\
10\,min & \textbf{0.79} & 922 \\
20\,min & \textbf{0.79} & 1843 \\
\bottomrule
\end{tabular}
\end{table}

\subsection{What the carried state holds}
\label{app:prime}

A state carried through a whole session could be doing either of two things: accumulating the
context of what has been typed, or adapting to the person typing it. The two predict opposite
results when the decoder is started from a state built somewhere else. If the state were adapting
to the typist, priming it on another session of the same user should help.

It does not (Table~\ref{tab:app-prime}). Priming on another session of the same user costs 0.82
CER, priming on a different user costs 3.66, and replaying the session's own first four minutes is
worth 0.06, inside the 0.17 that two decodes of one checkpoint differ by. No primed state beats a
cold start. The state carries context, and nothing here supports a claim that the decoder adapts
to its user.

\begin{table}[H]
\centering\small
\setlength{\tabcolsep}{6pt}
\caption{Starting the decoder from a state built elsewhere (test set, CER \%, mean over two
encoders; the direction is the same on both). The shipped system is 25.13
(Table~\ref{tab:main}).}
\label{tab:app-prime}
\begin{tabular}{@{}lrr@{}}
\toprule
Decoder started from & CER & \\
\midrule
cold, as shipped & 25.24 & \\
a replay of this session's first 4\,min & 25.30 & $+0.06$ \\
another session of the same user & 26.07 & \textcolor{loss}{$+0.82$} \\
another user's session & 28.90 & \textcolor{loss}{$+3.66$} \\
\bottomrule
\end{tabular}
\end{table}

\subsection{Causality certificate}
\label{app:causal}

A decoder that claims to be causal should be unaffected by anything that has not happened yet, and
that is directly testable. We fix a cut point in a session, then do one of three things to
everything after it: truncate the session there, replace it with noise, or shuffle it in time. We
then compare the emissions before the cut against the untouched run.

Every emission before every cut is bit-identical under all three interventions
(Table~\ref{tab:app-causal}). Argmax agreement is exactly 1 and the largest change in any log
probability is exactly 0, against thresholds of 0.999 and 0.05 fixed before the run.

The one place future information could enter is the running normaliser, whose first frames share a
single set of statistics. Shrinking that warm-up from 125 frames to 1 moves CER by $-0.09$,
$+0.04$, $-0.03$ and $-0.03$ on four encoders, below the 0.17 that two decodes of one checkpoint
differ by and inconsistent in sign.

\begin{table}[H]
\centering\small
\setlength{\tabcolsep}{6pt}
\caption{Emissions before a cut, against the same run with the future replaced. Two encoders and
14 session-and-cut combinations per intervention, cuts at 40\,s, 160\,s, 480\,s and 800\,s into the
session.}
\label{tab:app-causal}
\begin{tabular}{@{}lrrr@{}}
\toprule
Replacement after the cut & Cells & Argmax agreement & $\max|\Delta \log p|$ \\
\midrule
session truncated there & 28 & 1.000000 & 0.0 \\
noise & 28 & 1.000000 & 0.0 \\
the same frames, shuffled & 28 & 1.000000 & 0.0 \\
\bottomrule
\end{tabular}
\end{table}

\subsection{State vs Normalization}
\label{app:svsn}

\begin{table}[H]
\centering\small
\caption{What a reset destroys inside the encoder. Greedy CER (\%), mean $\pm$ s.d. over four
encoders on the 16 test sessions; the first row resets nothing and is the reference. The
interaction row is both minus the two single effects, so the two shares do not add up.}
\label{tab:reset-factorial}
\begin{tabular}{@{}lrr@{}}
\toprule
reset at the boundary & every 4\,s & every 32\,s \\
\midrule
nothing reset (CER) & $37.49$ & $37.50$ \\
recurrent state only & \textcolor{loss}{$+8.96 \pm 0.17$} & \textcolor{loss}{$+2.24 \pm 0.25$} \\
causal normaliser only & \textcolor{loss}{$+5.19 \pm 0.04$} & \textcolor{loss}{$+1.00 \pm 0.08$} \\
both & \textcolor{loss}{$+10.46 \pm 0.28$} & \textcolor{loss}{$+2.46 \pm 0.21$} \\
\bottomrule
\end{tabular}
\end{table}

\subsection{What each internal contributes}
\label{app:internals}

We remove the input-dependence of one internal at a time, leaving everything else untouched, and
see what the encoder costs (Table~\ref{tab:internals}). Two of the four carry the recurrence. The
two that are specific to this variant of Mamba cost almost nothing.

\begin{table}[H]
\centering\small
\setlength{\tabcolsep}{6pt}
\caption{Cost of removing one internal, greedy CER (\%) against a reference of 19.83 validation.}
\label{tab:internals}
\begin{tabular}{@{}llr@{}}
\toprule
Internal & What it does & Cost \\
\midrule
step size & how fast the state moves & \textcolor{loss}{it stops emitting} \\
decay & how long the state remembers & \textcolor{loss}{$+22.0$} \\
rotary clock & the state's phase & \textcolor{loss}{$+2.2$} \\
trapezoid split & how the step is discretised & \textcolor{loss}{$+0.4$} \\
\bottomrule
\end{tabular}
\end{table}

\subsection{Overlap between WikiText and the benchmark's prompts}
\label{app:overlap}

The benchmark's prose prompts are Wikipedia sentences and our corpus is drawn from WikiText-103,
which is also Wikipedia, so some overlap is expected and has to be quantified rather than assumed
away. Two things bound it. The two benchmark text files enter the corpus build only as a
decontamination blocklist, never as training text, and 921 generated lines were dropped by it. And
the residual overlap with the test prompts is small (Table~\ref{tab:app-overlap}): 1.13\% of test
prose words fall inside an eight-gram that also occurs in WikiText-103, falling to 0.38\% at
sixteen, and no word-list prompt overlaps at any length, which matters because word lists are about
half of the benchmark's lines.

The row that gives those figures their scale is the last one. The in-domain typing model, which we
report as a secondary result, is trained on the benchmark's own transcripts and overlaps the test
set six times more heavily, reproducing two per cent of test lines verbatim. Our corpus is not
free of Wikipedia, but it is an order of magnitude further from the test text than the in-domain
alternative, and it contains no benchmark transcript at all.

\begin{table}[H]
\centering\small
\setlength{\tabcolsep}{6pt}
\caption{Overlap with the benchmark's test prompts: share of prompt words lying inside an $n$-gram
that also occurs in the training text, and share of prompt lines reproduced verbatim.}
\label{tab:app-overlap}
\begin{tabular}{@{}lrrr@{}}
\toprule
Training text & 8-gram & 16-gram & Lines verbatim \\
\midrule
WikiText-103, prose prompts & 1.13 & 0.38 & 0.00 \\
WikiText-103, word-list prompts & 0.00 & 0.00 & 0.00 \\
the benchmark's own transcripts & 6.80 & 3.57 & 2.00 \\
\bottomrule
\end{tabular}
\end{table}

\subsection{What each part of the retokenizer contributes}
\label{app:sampler-ablate}

We remove one component of the retokenizer at a time, leaving the rest fitted as before, and
generate against the same held-out typists. Each component owns a statistic, and removing it moves
that statistic away from real typing (Table~\ref{tab:app-sampler-components}).

The per-typist layer and the real-word slips are the clearest. Without the first, the spread of
error rate across typists more than halves, and the corpus becomes a population of one average
typist; without the second, typos land on another real word half as often as they should, and those
are exactly the errors a spelling prior would let through. The correction process is the extreme
case: remove it and nothing survives uncorrected at all, against 4.85 uncorrected errors per
thousand keys in real typing. The count frailty and the burst process are smaller, at roughly two
and three times the seed band respectively, and we do not claim more for them than that.

\begin{table}[H]
\centering\small
\setlength{\tabcolsep}{6pt}
\caption{Each component against the statistic it accounts for, on held-out typists. The twenty-seed
band at fixed parameters is $\pm0.09$ for the typist spread, $\pm0.049$ for uncorrected errors and
$\pm0.0001$ for real-word errors.}
\label{tab:app-sampler-components}
\begin{tabular}{@{}llrrr@{}}
\toprule
Component & Statistic it accounts for & Real & Ours & Without it \\
\midrule
per-typist layer   & spread of error rate across typists & 6.38 & 3.77 & 1.65 \\
count frailty      & spread of error rate across typists & 6.38 & 3.77 & 3.57 \\
real-word slips    & typos landing on another real word  & 0.01329 & 0.01315 & 0.00799 \\
burst process      & errors never corrected, per 1{,}000 & 4.849 & 4.871 & 5.017 \\
correction process & errors never corrected, per 1{,}000 & 4.849 & 4.871 & 0.000 \\
\bottomrule
\end{tabular}
\end{table}

\end{document}